*Review*

# Particle Swarm Optimization: A survey of historical and recent developments with hybridization perspectives

**Saptarshi Sengupta \*, Sanchita Basak and Richard Alan Peters II**

Department of Electrical Engineering and Computer Science, Vanderbilt University, 2201 West End Ave, Nashville, TN 37235, USA; sanchita.basak@vanderbilt.edu (S.B.); alan.peters@vanderbilt.edu (R.A.P.)

**\*** Correspondence: saptarshi.sengupta@vanderbilt.edu; Tel.: +1 (615)-678-3419

**Abstract**: Particle Swarm Optimization (PSO) is a metaheuristic global optimization paradigm that has gained prominence in the last two decades due to its ease of application in unsupervised, complex multidimensional problems which cannot be solved using traditional deterministic algorithms. The canonical particle swarm optimizer is based on the flocking behavior and social co-operation of birds and fish schools and draws heavily from the evolutionary behavior of these organisms. This paper serves to provide a thorough survey of the PSO algorithm with special emphasis on the development, deployment and improvements of its most basic as well as some of the very recent state–of-the-art implementations. Concepts and directions on choosing the inertia weight, constriction factor, cognition and social weights and perspectives on convergence, parallelization, elitism, niching and discrete optimization as well as neighborhood topologies are outlined. Hybridization attempts with other evolutionary and swarm paradigms in selected applications are covered and an up-to-date review is put forward for the interested reader.

**Keywords:** Particle Swarm Optimization; Swarm Intelligence; Evolutionary Computation; Intelligent Agents; Optimization; Hybrid Algorithms; Heuristic Search; Approximate Algorithms; Robotics and Autonomous Systems; Applications of PSO

**1. Introduction**

The last two decades has seen unprecedented development in the field of Computational Intelligence with the advent of the GPU and the introduction of several powerful optimization algorithms that make little or no assumption about the nature of the problem. Particle Swarm Optimization (PSO) is one among many such techniques and has been widely used in treating ill-structured continuous/discrete, constrained as well as unconstrained function optimization problems [1]. Much like popular Evolutionary Computing paradigms such as Genetic Algorithms [2] and Differential Evolution [3], the inner workings of the PSO make sufficient use of probabilistic transition rules to make parallel searches of the solution hyperspace without explicit assumption of derivative information. The underlying physical model upon which the transition rules are based is one of emergent collective behavior arising out of social interaction of flocks of birds and schools of fish. Since its inception in 1995, PSO has found use in an ever-increasing array of complex, real-world optimization problems where conventional approaches either fail or render limited usefulness. Its intuitively simple representation and relatively low number of adjustable parameters make it a popular choice for many problems which require approximate solutions up to a certain degree. There are however, several major shortcomings of the basic PSO that introduce failure modes such as stagnation and convergence to local optima which has led to extensive studies (such as [4-5]) aimed at mitigation and resolution of the same. In this review, the foundations and frontiers of advances in

PSO have been reported with a thrust on significant developments over the last decade. The remainder of the paper is organized sequentially as follows: Section 2 provides a historical overview and motivation for the Particle Swarm Optimization algorithm, Section 3 outlines the working mechanism of PSO and Section 4 details perspectives on historical and recent advances along with a broad survey of hybridization approaches with other well-known evolutionary algorithms. Section 5 reviews niche formation and multi-objective optimization discussing formation of niches in PSO and niching in dynamic environments. This is followed in Section 6 by an informative review of the applications of PSO in discrete optimization problems and in Section 7 by notes on ensemble optimizers. Section 8 presents notes on benchmark solution quality and performance comparison practices and finally, Section 9 outlines future directions and concludes the paper.

## 2. The Particle Swarm Optimization: Historical Overview

Agents in a natural computing paradigm are decentralized entities with generally no perception of the high-level goal in pursuit yet can model complex real-world systems. This is made possible through several low-level goals which when met facilitate meaningful collective behavior arising from these seemingly unintelligent and non-influential singular agents. An early motivation can be traced from Reeves' introduction of *particle systems* in the context of modeling natural objects such as fire, clouds and water in computer based animations while at Lucasfilm Ltd (1983) [6]. In the course of development, agents or 'particles' are generated, undergo transformations in form and move around in the modeling environment and eventually are rejected or 'die'. Reeves concluded that such a model is able to represent the dynamics and form of natural environments which were rendered infeasible using classical surface-based representations [6]. Subsequent work by Reynolds in the *Boid Model* (1986) established simple rules that increased autonomy of particle behavior and laid down simple low-level rules that *boids* (bird-oid objects) or particles could obey to give rise to emergent behavior [7]. The complexity of the Boids Model is thus a direct derivative of the simple interactions between the individual particles. Reynolds formulated three distinct rules of flocking for a particle to follow: separation, alignment and cohesion. While the separation principle allows particles to move away from each other to avoid crowding, the alignment and cohesion principles necessitate directional updates to move towards the average heading and position of nearby flock members respectively. The inherent nonlinearity of the boids render *chaotic behavior* in the emergent group dynamics whereas the negative feedback introduced by the simple, low level rules effect in *ordered behavior*. The case where each boid knows the whereabouts of every other boid has $O(n^2)$ complexity making it computationally infeasible. However, Reynolds propositioned a neighborhood model with information exchange among boids in a general vicinity, thereby reducing the complexity to $O(n)$ and speeding up the algorithmic implementation. The Particle Swarm Optimization algorithm was formally introduced in 1995 by Eberhart and Kennedy through an extension of Reynold's work. By incorporating local information exchange through nearest neighbor velocity matching, the flock or *swarm* prematurely converged in a unanimous fashion. Hence, a random perturbation or *craziness* was introduced in the velocities of the particles leading to sufficient variation and subsequent lifelike dynamics of the swarm. Both these parameters were later eliminated as the flock seemed to converge onto *attractors* equally well without them. The paradigm thus ended up with a population of agents which were more in conformity with the dynamics of a swarm than a flock.

## 3. Working Mechanism of the canonical PSO

The PSO algorithm employs a swarm of particles which traverse a multidimensional search space to seek out optima. Each particle is a potential solution and is influenced by experiences of its neighbors as well as itself. Let $x_i(t)$ be the position in the search space of the *i*-th particle at time step *t*. The initial velocity of a particle is regulated by incrementing it in the positive or negative direction contingent on the current position being less than the best position and vice-versa (Shi and Eberhart, 1998) [8].

$$vx[\ ][\ ] = vx[\ ][\ ] + 2*rand()*(pBest[\ ][\ ] - presentx[\ ][\ ]) + 2*rand() \\ *(pBest[\ ][gbest] - presentx[\ ][\ ]) \quad (1)$$

The random number generator was originally multiplied by 2 in [1] so that particles could have an overshoot across the target in the search space half of the time. These values of the constants, known as the cognition and social acceleration co-efficient were found to effect superior performance than previous versions. Since its introduction in 1995, the PSO algorithm has undergone numerous improvements and extensions aimed at guaranteeing convergence, preserving and improving diversity as well as offsetting the inherent shortcomings by hybridizing with parallel EC paradigms.

**4. Perspectives on Development**

*4.1. Inertia Weight*

The initialization of particles is critical in visiting optima when the initial velocity is zero. This is because the pBest and gBest attractors help intelligently search the neighborhood of the initial kernel but do not facilitate exploration of new regions in the search space. The velocity of the swarm helps attain this purpose, however suitable clamps on the velocity are needed to ensure the swarm does not diverge. Proper selection of the maximum velocity $v_{max}$ is important to maintain control: a large $v_{max}$ introduces the possibility of global exploration whereas a small value implies a local, intensive search. Shi and Eberhart suggested an *'inertia weight'* $\omega$ which is used as a control parameter for the swarm velocity [8], thereby making possible the modulation of the swarm's momentum using constant, linear time-varying or even non-linear temporal dependencies [9]. However, the inertia weight could not fully do away with the necessity for velocity clamping [10]. To guarantee convergent behavior and to come to a balance between exploitation and exploration the value of the inertia weight must be chosen with care. An inertia weight equal or greater than one implies the swarm velocity increases over time towards the maximum velocity $v_{max}$. Two things happen when the swarm velocity accelerates rapidly towards $v_{max}$: particles cannot change their heading to fall back towards promising regions and eventually the swarm diverges. On the other hand, an inertia weight less than one reduces the acceleration of the swarm until it eventually becomes a function of only the acceleration factors. The exploratory ability of the swarm suffers as the inertia goes down, making sudden changes in heading possible as social and cognitive factors increasingly control the position updates. Early works on the inertia weight used a constant value throughout the course of the iterations but subsequent contributions accommodated the use the dynamically changing values. The de facto approach seemed to be to use a large initial value of $\omega$ to help in global exploration followed by a gradual decrease to hone in on promising areas towards the latter part of the search process.

Efforts to dynamically change the inertia weight can be organized in the following categorizations:

4.1.1. Random Selection (RS)

In each iteration, a different inertia weight is selected, possibly drawn from an underlying distribution with a mean and standard deviation of choice. However, care should be taken to ensure convergent behavior of the swarm.

4.1.2. Linear Time Varying (LTV)

Usually, the implementation of this kind decreases the value of $\omega$ from a preset high value of $\omega_{max}$ to a low of $\omega_{min}$. Standard convention is to take $\omega_{max}$ and $\omega_{min}$ as 0.9 and 0.4. The LTV inertia weight can be expressed as [11-12]:

$$\omega_t = (\omega_{max} - \omega_{min})\frac{(t_{max} - t)}{t_{max}} + \omega_{max} \quad (2)$$

where $t_{max}$ is the number of iterations, $t$ is the current iteration and $\omega_t$ is the value of the inertia weight in the $t$-th iteration.

There are some implementations that look at the effects of increasing the inertia weight from an initial low value to a high value, the interested reader should refer to [13-14].

4.1.3. Non-Linear Time Varying (NLTV)

As with LTV inertia weights, NLTV inertia weights too, tend to fall off from an initial high value at the start of the optimization process. Nonlinear decrements allow more time to fall off towards the lower end of the dynamic range, thereby enhancing local search or exploitation. Naka et al. [15] proposed the following nonlinear time varying inertia weight:

$$\omega_{t+1} = (\omega_t - 0.4)\frac{(t_{max} - t)}{t_{max} + 0.4} \qquad (3)$$

where $\omega_{t=0} = 0.9$ is the initial choice of $\omega$. Clerc introduced the concept of relative improvement of the swarm in developing an adaptive inertia weight [16]. The change in the inertia of the swarm is in proportion to the relative improvement of the swarm. The relative improvement $\kappa^i_t$ is estimated by:

$$\kappa^i_t = \frac{f(lbest^t_t) - f(x^t_t)}{f(lbest^t_t) + f(x^t_t)} \qquad (4)$$

Clerc's updated inertia weight can be expressed as:

$$\omega_{t+1} = \omega_0 + (\omega_{t_{max}} - \omega_0)\frac{e^{m_i(t)} - 1}{e^{m_i(t)} + 1} \qquad (5)$$

where $\omega_{tmax}=0.5$ and $\omega_0<1$. Each particle has a unique inertia depending on its distance from the local best position.

4.1.4. Fuzzy Adaptive (FA)

Using fuzzy sets and membership rules, $\omega$ can be dynamically updated as in [17]. The change in inertia may be computed using the fitness of the gBest particle as well as that of the current value of $\omega$. The change is implemented through the use of a set of fuzzy rules as in [17-18].

The choice of $\omega$ is thus dependent on the optimization problem in hand, specifically on the nature of the search space.

*4.2. Constriction Factor*

Clerc demonstrated that to ensure optimal trade-off between exploration and exploitation, the use of a constriction coefficient $\chi$ may be necessary [19-20]. The constriction co-efficient was developed from eigenvalue analyses of computational swarm dynamics in [19]. The velocity update equation changes to:

$$vx[\,][\,] = \chi(vx[\,][\,] + \Omega_1 * rand(\,) * (pBest[\,][\,] - presentx[\,][\,]) + \Omega_2 * rand(\,) \\ * (pBest[\,][gbest\,] - presentx[\,][\,])) \qquad (6)$$

where $\chi$ was shown to be:

$$\chi = \frac{2\nu}{\left|2 - \Omega - \sqrt{\Omega(\Omega - 4)}\right|} \qquad (7)$$

$$\Omega = \Omega_1 + \Omega_2 \qquad (8)$$

$\Omega_1$ and $\Omega_2$ can be split into products of social and cognitive acceleration coefficients $c_1$ and $c_2$ times random noise $r_1$ and $r_2$. Under the operating constraint that $\Omega \geq 4$ and $\nu \in [0,1]$, swarm convergence is guaranteed with particles decelerating as iteration count increases. The parameter $\nu$ controls the local or global search scope of the swarm. For example, when $\nu$ is set close to 1, particles traverse the search space with a predominant emphasis on exploration. This leads to slow convergence and a high degree of accuracy in finding the optimum solution, as opposed to when $\nu$ is close to zero in which case the convergence is fast but the solution quality may vary vastly. This approach of constricting the velocities is equivalent in significance to the inertia weight variation, given its impact on determining solution quality across neighborhoods in the search space. Empirical studies in [21] demonstrated that faster convergence rates are achieved when velocity constriction is used in conjunction with clamping.

*4.3. Cognition and Social Velocity Models of the Swarm*

One of the earliest studies on the effect of different attractors on the swarm trajectory update was undertaken by Kennedy in 1997 [22]. The cognition model considers only the cognitive component of the canonical PSO in Equation (1).

$$v_{t+1}[\ ][\ ] = (v_t[\ ][\ ] + C_1 * rand(\ ) * (pBest[\ ][\ ] - presentx[\ ][\ ]) \tag{9}$$

The cognition model performs a local search in the region where the swarm members are initialized and tends to report suboptimal solutions if the acceleration component and upper bounds on velocity are small. Due to its weak exploratory ability, it is also slow in convergence. This was reported by Kennedy [22] and subsequently the subpar performance of the model was confirmed by the works of Carlisle and Dozier [23]. The social model, on the other hand, considers only the social component.

$$v_{t+1}[\ ][\ ] = (v_t[\ ][\ ] + C_1 * rand(\ ) * (pBest[\ ][gBest] - presentx[\ ][\ ]) \tag{10}$$

In this model, the particles are attracted towards the global best in the feasible neighborhood and converge faster with predominantly exploratory behavior. This was reported by Kennedy [22] and confirmed by Carlisle and Dozier [23].

*4.4. Cognitive and Social Acceleration Coefficients*

The acceleration coefficients $C_1$ and $C_2$ when multiplied with random vectors $r_1$ and $r_2$ render controllable stochastic influences on the velocity of the swarm. $C_1$ and $C_2$, simply put, are weights that capture how much a particle should weigh moving towards its cognitive attractor (pBest) or its social attractor (gBest). Exchange of information between particles mean they are inherently co-operative, thus implying that an unbiased choice of the acceleration coefficients would make them equal. For case specific implementations, one may set $C_1=0$ or $C_2=0$ with the consequence being that individual particles will rely solely on their own knowledge or that individual particles will only rely on the knowledge of the best particle in the entire swarm. It is obvious that multimodal problems containing multiple promising regions will benefit from a balance between social and cognitive components of acceleration.

Stacey et al. used mutation functions in formulating acceleration coefficients and by keeping the step size of mutation equal to $v_{max}$, improvements were noticed over the general implementation (MPSO-TVAC) [24]. Jie et al. introduced new Metropolis coefficients in PSO, leading to better efficiency and stability [25]. It hybridizes Particle Swarm Optimization with Simulated Annealing [26] and reduces runtime as well as the number of iterations.

4.4.1. Choice of Values

In general, the values of $C_1$ and $C_2$ are kept constant. An empirically found optimum pair seems to be 2.05 for each of $C_1$ and $C_2$ and significant departures or incorrect initializations lead to divergent behavior. Ratnaweera et al. suggested that $C_1$ should be decreased linearly over time, whereas $C_2$ should be increased linearly [27]. Clerc's fuzzy acceleration reports improvements using swarm diversity and the ongoing iteration by adaptively refining coefficient values [16].

*4.5. Topologies*

The topology of the swarm of particles establishes a measure of the degree of connectivity of its members to the others. It essentially describes a subset of particles with whom a particle can initiate information exchange [28]. The original PSO outlined two topologies that led to two variants of the algorithm: *lBest PSO* and *gBest PSO*. The *lBest* variant associates a fraction of the total number of particles in the neighborhood of any particular particle. This structure leads to multiple best particles, one in each neighborhood and consequently the velocity update equation of the PSO has multiple social attractors. Under such circumstances, the swarm is not attracted towards any single global best rather a combination of sub-swarm bests. This brings down the convergence speed but significantly increases the chance of finding global optima. In the *gBest* variant, all particles simultaneously influence the social component of the velocity update in the swarm, thereby leading to an increased convergence speed and a potential stagnation to local optima if the true global optima is not where the best particle of the neighborhood is.

There have been some fundamental contributions to the development of PSO topologies over the last two decades [29-31]. A host of PSO topologies have risen out of these efforts, most notably the Random Topology PSO, The Von-Neumann Topology PSO, The Star Topology PSO and the Toroidal Topology PSO. In [31], Mendes et al. studied several different sociometry with a population size of 20 where they quantified the effect of including the past experiences of an individual by implementing with and without, the particle of interest. Interested readers can also refer to the recent work by Liu et al to gain an understanding about topology selection in PSO driven optimization environments [32].

*4.6. Analysis of Convergence*

In this section, the underlying constraints for convergence of the swarm to an equilibrium point are reviewed. Van den Bergh and Engelbrecht as well as Trelea noted that the trajectory of an individual particle would converge contingent upon meeting the following condition [33-35]:

$$1 > \omega > \frac{\Omega_1 + \Omega_2}{2} - 1 \geq 0 \tag{11}$$

The above relation can be simplified by replacing the stochastic factors with the acceleration coefficients $C_1$ and $C_2$ such that when $C_1$ and $C_2$ are chosen to satisfy the condition in Eq. (12), the swarm converges.

$$1 > \omega > \frac{C_1 + C_2}{2} - 1 \geq 0 \tag{12}$$

Studies in [19, 34-35] also lead to the implication that a particle may converge to a single point $X'$ which is a stochastic attractor with pBest and gBest being ends of two diagonals. This point may not be an optimum and particles may prematurely converge to it.

$$X' = \frac{\Omega_1 pBest + \Omega_2 gBest}{\Omega_1 + \Omega_2} \tag{13}$$

*4.7. Velocity and Position Update Equations of the Standard PSO*

The following equations describe the velocity and position update mechanisms in a standard PSO algorithm:

$$v_{ij}(t+1) = \omega v_{ij}(t) + r_1(t)C_1\left(pbest_{ij}(t) - x_{ij}(t)\right) + r_2(t)C_2\left(gbest(t) - x_{ij}(t)\right) \tag{14}$$

$$x_{ij}(t+1) = x_{ij}(t) + v_{ij}(t+1) \tag{15}$$

$r_1$ and $r_2$ are independent and identically distributed random numbers whereas $C_1$ and $C_2$ are the cognition and social acceleration coefficients. $x_{ij}$, $v_{ij}$ are position coordinates and velocity of the $i_{th}$ agent in the $j_{th}$ dimension. $pbest_{ij}(t)$ and $gbest(t)$ represent the personal and global best locations in the *t-th* iteration. The first term in the right-hand side of eq. (14) makes use of $\omega$, which is the inertia weight and the next two terms are excitations towards promising regions in the search space as reported by the personal and global best locations. The personal best replacement procedure assuming a function minimization objective is discussed in eq. (16). The global best *gbest(t)* is the minimum cost bearing element of the temporal set of personal bests *pbest_i(t)* of all particles over all iterations.

$$\text{if } cost(x_i(t+1)) < cost(pbest_i(t)) \Rightarrow pbest_i(t+1) = x_i(t+1) \tag{16}$$

$$\text{else } pbest_i(t+1) = pbest_i(t)$$

*4.8. Survey of Hybridization Approaches*

A hybridized PSO implementation integrates the inherent social, co-operative character of the algorithm with tested optimization strategies arising out of distinctly different traditional or evolutionary paradigms towards achieving the central goal of intelligent exploration-exploitation. This is particularly helpful in offsetting weaknesses in the underlying algorithms and distributing the randomness in a guided way. The literature on hybrid PSO algorithms is quite rich and growing by the day. In this section, some of the most notable works as well as a few recent approaches have been outlined.

4.8.1. Hybridization of PSO using Genetic Algorithms (GA)

Popular approaches in hybridizing GA and PSO involve using the two approaches sequentially or in parallel or by using GA operators such as selection, mutation and reproduction within the PSO framework. Authors in [36] used one algorithm until stopping criterion is reached to use the final solution in the other algorithm for fine tuning. How the stopping criterion is chosen varies. They used a switching method between the algorithms when one algorithm fails to improve upon past results over a chosen number of iterations. In [37] the first algorithm is terminated once a specified number of iterations has been exceeded. The best particles from the first algorithm populate the particle pool in the second algorithm and the empty positions are filled using random generations. This preserves the diversity of the otherwise similar performing population at the end of the first phase. Authors in [37] put forth the idea of exchanging fittest particles between GA and PSO, running in parallel for a fixed number of iterations.

In Yang et al.'s work on PSO-GA based hybrid evolutionary algorithm (HEA) [38], the evolution strategy of particles employs a two-phase mechanism where the evolution process is accelerated by using PSO and diversity is maintained by using GA. The authors used this method to optimize three unconstrained and three constrained problems. Li et al. used mechanisms such as nonlinear ranking selection to generate offspring from parents in a two-stage hybrid GA-PSO where each stage is separately accomplished using GA and PSO [39]. Valdez et al. proposed a fuzzy approach in testing PSO-GA hybridization [40]. Simple fuzzy rules were used to determine whether to consider GA or PSO particles and change their parameters or to take a decision. Ghamisi and Benediktsson [55] introduced a feature selection methodology by hybridizing GA and PSO. This method was tested on the Indian Pines hyperspectral dataset as well as for road detection purposes. The accuracy of a Support Vector Machine (SVM) classifier on validation samples was set as the fitness score. The method could select the most informative features within an acceptable processing time automatically and did not require the users to set the number of desired features beforehand.

Benvidi et al. [56] used a hybrid GA-PSO algorithm to simultaneously quantify four commonly used food colorants containing tartrazine, sunset yellow, ponceau 4R and methyl orange, without prior chemical separation. Results indicated the designed model accurately determined concentrations in real as well as synthetic samples. From observations the introduced method emerged as a powerful tool to estimate the concentration of food colorants with a high degree of overlap using nonlinear artificial neural network. Yu et al. used a hybrid PSO-GA to estimate energy demand of China in [57] whereas Moussa and Azar introduced a hybrid algorithm to classify software modules as fault-prone or not using object-oriented metrics in [58]. Nik et al. used GA-PSO, PSO-GA and a collection of other hybridization approaches to optimize surveyed asphalt pavement inspection units in massive networks [59]. Premlatha and Natarajan [52] proposed a discrete version of PSO with embedded GA operators for clustering purposes. The GA operator initiates reproduction when particles stagnate. This version of the hybrid algorithm was named DPSO with mutation-crossover.

In [53] Abdel-Kader proposed a GAI-PSO hybrid algorithm for k-means clustering. The exploration ability of the algorithm was used first to find an initial kernel of solutions containing cluster centroids which was subsequently used by the k-means in a local search. For treating constrained optimization problems, Garg used a PSO to operate in the direction of improving the vector while using GA to update decision vectors [60]. In [61], Zhang et al. carried out experimental investigations to optimize the performance of a four-cylinder, turbocharged, direct-injection diesel engine. A hybrid PSO and GA method with a small population was tested to optimize five operating parameters, including EGR rate, pilot timing, pilot ratio, main injection timing, and injection pressure. Results demonstrated significant speed-up and superior optimization as compared to GA. Li et al. developed a mathematical model of the heliostat field and optimized it using PSO-GA to determine the highest potential daily energy collection (DEC) in [62]. Results indicated that DEC during the spring equinox, summer solstice, autumnal equinox and winter solstice increased approximately by $1.1 \times 10^5$ MJ, $1.8 \times 10^5$ MJ, $1.2 \times 10^5$ MJ and $0.9 \times 10^5$ MJ, respectively.

A brief listing of some of the important hybrid algorithms using GA and PSO are given below in Table 1.

**Table 1.** A collection of hybridized GA-PSO algorithms.

| Author/s: | Year | Algorithm | Area of Application |
|---|---|---|---|
| Robinson et al. [36] | 2002 | GA-PSO, PSO-GA | Engineering design optimization |
| Krink and Løvbjerg [41] | 2002 | Life Cycle Model | Unconstrained global optimization |
| Conradie et al. [42] | 2002 | SMNE | Neural Networks |
| Grimaldi et al. [43] | 2004 | GSO | Electromagnetic Application |
| Juang [44] | 2004 | GA-PSO | Network Design |
| Settles and Soule [45] | 2005 | Breeding Swarm | Unconstrained Global Optimization |
| Jian and Chen [46] | 2006 | PSO-RDL | Unconstrained Global Optimization |
| Esmin et al. [47] | 2006 | HPSOM | Unconstrained Global Optimization |
| Kim [48] | 2006 | GA-PSO | Unconstrained Global Optimization |
| Mohammadi and Jazaeri [49] | 2007 | PSO-GA | Power Systems |
| Gandelli et al. [50] | 2007 | GSO | Unconstrained Global Optimization |
| Yang et al. [38] | 2007 | HEA | Constrained and Unconstrained Global Optimization |
| Kao and Zahara [51] | 2008 | GA-PSO | Unconstrained Global Optimization |
| Premlatha and Natrajan [52] | 2009 | DPSO-mutation-crossover | Document Clustering |
| Abdel Kader [53] | 2010 | GAI-PSO | Data Clustering |
| Kuo and Hong [54] | 2013 | HGP1, HGP2 | Investment Portfolio Optimization |
| Ghamisi and Benedictsson [55] | 2015 | GA-PSO | Feature Selection |
| Benvidi et al [56] | 2016 | GA-PSO | Spectrophotometric determination of synthetic colorants |
| Yu et al. [57] | 2011 | GA-PSO | Estimation of Energy Demand |
| Moussa and Azar [58] | 2017 | PSO-GA | Classification |
| Nik, Nejad and Zakeri [59] | 2016 | GA-PSO, PSO-GA | Optimization of Surveyed Asphalt Pavement Inspection Unit |
| Garg [60] | 2015 | GA-PSO | Constrained Optimization |
| Zhang et al. [61] | 2015 | PSO-GA | Biodiesel Engine Performance Optimization |
| Li et al. [62] | 2018 | PSO-GA | Optimization of a heliostat field layout |

### 4.8.2. Hybridization of PSO using Differential Evolution (DE)

Differential Evolution (DE) by Price and Storn [63] is a very popular and effective metaheuristic for solving global optimization problems. Several approaches of hybridizing DE with PSO exist in the literature, some of which are elaborated in what follows.

Hendtlass [64] introduced a combination of particle swarm and differential evolution algorithm (SDEA) and tested it on a graduated set of trial problems. The SDEA algorithm works the same way as a particle swarm one, except that DE is run intermittently to move particles from worse performing areas to better ones. Experiments on a set of four benchmark problems, viz. The Goldstein-Price Function, the six hump camel back function, the Timbo2 Function and the n-dimensional 3 Potholes Function showed improvements in performance. It was noted that the new algorithm required more fitness evaluations and that it would be feasible to use the component swarm based algorithm for problems with computationally heavy fitness functions. Zhang and Xie [65] introduced another variant of a hybrid DE-PSO. Their strategy employed different operations at random, rather than a combination of both at the same time. Results on benchmark problems indicated better performance than PSO or DE alone. Talbi and Batouche [66] used DEPSO to approach the multimodal rigid-body image registration problem by finding the optimal transformation which superimposed two images by maximization of mutual information. Hao et al. [67] used selective updates for the particles' positions by using partly a DE approach, partly a PSO approach and tested it on a suite of benchmark problems.

Das et al. scrapped the cognitive component of the velocity update equation in PSO and replaced it with a weighted difference vector of positions of any two different particles chosen randomly from the population [68]. The modified algorithm was used to optimize well-known benchmarks as well as constrained optimization problems. The authors demonstrated the superiority of the proposed method, achieved through a synergism between the underlying popular multi-agent search processes: the PSO and DE. Luitel and Venayagamoorthy [69] used a DEPSO optimizer to design linear phase Finite Impulse Response (FIR) filters. Two different fitness functions: one based on passband and stopband ripple, the other on MSE between desired and practical results were considered. While promising results were obtained with respect to performance and convergence time, it was noted that the DEPSO algorithm could also be applied to the personal best position, instead of the global best. Vaisakh et al. [70] came up with a DEPSO algorithm to achieve optimal reactive power dispatch with reduced power and enhanced voltage stability. The IEEE-30 bus test system is used to illustrate its effectiveness and results confirm the superiority of the algorithm proposed. Huang et al. [71] studied the back analysis of mechanics parameters using DEPSO-ParallelFEM - a hybrid method using the advantages of DE fused with PSO and Finite Element Method (FEM). The DEPSO algorithm enhances the ability to escape local minima and the FEM increases computational efficiency and precision through involvement of Cluster of Workstation (COW), MPI (Message Passing Interface), Domain Decomposition Method (DDM) [72-73] and Object-oriented Programming (OOP) [74-75]. A computational example supports the claim that it is an efficient method to estimate and back analyze the mechanics parameters of systems.

Xu et al. [76] applied their proposed variant of DEPSO on data clustering problems. Empirical results obtained on synthetic and real datasets showed that DEPSO achieved faster performance than when either of PSO or DE is used alone. Xiao and Zuo [77] used a multi-population strategy to diversify the population and employ every sub-population to a different peak, subsequently using a hybrid DEPSO operator to find the optima in each. Tests on the Moving Peaks Benchmark (MPB) problem resulted in significantly better average offline error than competitor techniques. Junfei et al. [78] used DEPSO for mobile robot localization purposes whereas Sahu et al. [79] proposed a new fuzzy Proportional–Integral Derivative (PID) controller for automatic generation control of interconnected power systems. Seyedmahmoudian et al. [80] used DEPSO to detect maximum power point under partial shading conditions. The proposed technique worked well in achieving the Global Maximum Power Point (GMPP): simulation and experimental results verified this under different partial shading conditions and as such its reliability in tracking the global optima was established. Gomes and Saraiva [81] described a hybrid evolutionary tool to solve the Transmission Expansion

Planning problem. The procedure is phased out in two parts: first equipment candidates are selected using a Constructive Heuristic Algorithm and second, a DEPSO optimizer is used for final planning. A case study based on the IEEE 24-Bus Reliability Test System using the DEPSO approach yielded solutions of acceptable quality with low computational effort.

Boonserm and Sitjongsataporn [82] put together DE, PSO and Artificial Bee Colony (ABC) [129] coupled with self-adjustment weights determined using a sigmoidal membership function. DE helped eliminate the chance of premature convergence and PSO sped up the optimization process. The inherent ABC operators helped avoid suboptimal solutions by looking for new regions when fitness did not improve. A comparative analysis of DE, PSO, ABC and the proposed DEPSO-Scout over benchmark functions such as Rosenbrock, Rastrigin and Ackley was performed to support the claim that the new metaheuristic performed better than the component paradigms viz. PSO, DE and ABC.

A brief listing of some of the important hybrid algorithms using SA and PSO are given below in Table 2.

Table 2. A collection of hybridized DE-PSO algorithms.

| Author/s: | Year | Algorithm | Area of Application |
| --- | --- | --- | --- |
| Hendtlass [64] | 2001 | SDEA | Unconstrained Global Optimization |
| Zhang and Xie [65] | 2003 | DEPSO | Unconstrained Global Optimization |
| Talbi and Batouche [66] | 2004 | DEPSO | Rigid-body Multimodal Image Registration |
| Hao et al. [67] | 2007 | DEPSO | Unconstrained Global Optimization |
| Das et al. [68] | 2008 | PSO-DV | Design Optimization |
| Luitel and Venayagamoorthy [69] | 2008 | DEPSO | Linear Phase FIR Filter Design |
| Vaisakh et al. [70] | 2009 | DEPSO | Power Dispatch |
| Huang et al. [71] | 2009 | DEPSO-ParallelFEM | Back Analysis of Mechanics Parameters |
| Xu et al. [76] | 2010 | DEPSO | Clustering |
| Xiao and Zuo [77] | 2012 | Multi-DEPSO | Dynamic Optimization |
| Junfei et al. [78] | 2013 | DEPSO | Mobile Robot Localization |
| Sahu et al. [79] | 2014 | DEPSO | PID Controller |
| Seyedmahmoudian et al. [80] | 2015 | DEPSO | Photovoltaic Power Generation |
| Gomes and Saraiva [81] | 2016 | DEPSO | Transmission Expansion Planning |
| Boonserm and Sitjongsataporn [82] | 2017 | DEPSO-Scout | Numerical Optimization |

4.8.3 Hybridization of PSO using Simulated Annealing (SA)

Zhao et al. [83] put forward an activity network based multi-objective partner selection model and applied a new heuristic based on PSO and SA to solve the multi-objective problem. Yang et al. [84] produced one of the early works on PSO-SA hybrids in 2006 which detailed the embedding of SA in the PSO operation. They noted the efficient performance of the method on a suite of benchmark functions commonly used in the Evolutionary Computing (EC) literature. Gao et al. [85] trained a Radial Basis Function Neural Network (RBF-NN) using a hybrid PSO with chaotic search and simulated annealing. The component algorithms can learn from each other and mutually offset weak performances. Benchmark function optimization and classification results for datasets from the UCI Machine Learning Repository [86] demonstrated the efficiency of the proposed method. Chu et al. [87] developed an Adaptive Simulated Annealing-Parallel Particle Swarm Optimization (ASA-PPSO). ASA-PPSO uses standard initialization and evolution characteristics of the PSO and uses a greedy approach to replace the memory of best solutions. However, it also infuses an 'infix' condition which checks the latest two global best solutions and when triggered, applies an SA operator on some recommended particles or on all. Experimental analyses on benchmark functions established the usefulness of the proposed method.

Sadati et al. [88] formulated the Under-Voltage Load Shedding (UVLS) problem using the idea of static voltage stability margin and its sensitivity at the maximum loading point or the collapse point. The PSO-B-SA proposed by the authors was implemented in the UVLS scheme on the IEEE 14 and 18 bus test systems and considers both technical and economic aspects of each load. The proposed algorithm can reach optimum solutions in minimum runs as compared to the other competitive techniques in [88], thereby making it suitable for application in power systems which require an approximate solution within a finite time bound. Ma et al. [97] approached the NP-hard Job-shop Scheduling Problem using a hybrid PSO with SA operator as did Ge et al. in [89], Zhang et al. in [93] and Song et al. in [90]. A hybrid discrete PSO-SA algorithm was proposed by Dong et al. [91] to find optimal elimination orderings for Bayesian networks. Shieh et al. [92] devised a hybrid SA-PSO approach to solve combinatorial and nonlinear optimization problems. Idoumghar et al. [94] hybridized Simulated Annealing with Particle Swarm Optimization (HPSO-SA) and proposed two versions: a sequential and a distributed implementation. Using the strong local search ability of SA and the global search capacity of PSO, the authors tested out HPSO-SA on a set of 10 multimodal benchmark functions noting significant improvements. The sequential and distributed approaches are used to minimize energy consumption in embedded systems memories. Savings in terms of energy as well as execution time were noted. Tajbaksh et al. [95] proposed the application of a hybrid PSO-SA to solve the Traveling Tournament Problem.

Niknam et al. [96] made use of a proposed PSO-SA to solve the Dynamic Optical Power Flow Problem (DOPF) with prohibited zones, valve-point effects and ramp-rate constraints taken into consideration. The IEEE 30-bus test system was used to show the effectiveness of the PSO-SA in searching the possible solutions to the highly nonlinear and nonconvex DOPF problem. Sudibyo et al. [98] used SA-PSO for controlling temperatures of the trays in Methyl tert-Butyl Ether (MTBE) reactive distillation in a Non-Linear Model Predictive Control (NMPC) problem and noted the efficiency of the algorithm in finding the optima as a result of hybridization. Wang and Sun [99] applied a hybrid SA-PSO to the K-Means clustering problem.

Javidrad and Nazari [100] recently contributed a hybrid PSO-SA wherein SA contributes in updating the global best particle just when PSO does not show improvements in the performance of the global best particle, which may occur several times during the iteration cycles. The algorithm uses PSO in its initial phase to determine the global best and when there is no change in the global best in any particular cycle, passes the information on to the SA phase which iterates until a rejection takes place using the Metropolis criterion [101]. The new information about the best solution is then passed back to the PSO phase which again initiates search with the obtained information as the new global best. This process of sharing is sustained until convergence criteria are satisfied. Li et al. [102] introduced an efficient energy management scheme in order to increase the fuel efficiency of a Plug-In Hybrid Electric Vehicle (PHEV).

A brief listing of some of the important hybrid algorithms using SA and PSO are given below in Table 3.

**Table 3.** A collection of hybridized SA-PSO algorithms.

| Author/s: | Year | Algorithm | Area of Application |
|---|---|---|---|
| Zhao et al. [83] | 2005 | HPSO | Partner Selection for Virtual Enterprise |
| Yang et al. [84] | 2006 | PSOSA | Global Optimization |
| Gao et al. [85] | 2006 | HPSO | Optimizing Radial Basis Function |
| Chu et al. [87] | 2006 | ASA-PPSO | Global Optimization |
| Sadati et al. [88] | 2007 | PSO-B-SA | Under-voltage Load Shedding Problem |
| Ge et al. [89] | 2007 | Hybrid PSO with SA operator | Job-shop Scheduling |

| Song et al. [90] | 2008 | Hybrid PSO with SA operator | Job-shop Scheduling |
|---|---|---|---|
| Dong et al. [91] | 2010 | PSO-SA | Bayesian Networks |
| Shieh et al. [92] | 2011 | SA-PSO | Global Optimization |
| Zhang et al. [93] | 2011 | Hybrid PSO with SA operator | Job-shop Scheduling |
| Idoumghar et al. [94] | 2011 | HPSO-SA | Embedded Systems |
| Tajbaksh et al. [95] | 2012 | PSO-SA | Traveling Tournament Problem |
| Niknam et al. [96] | 2013 | SA-PSO | Dynamic Optical Power Flow |
| Ma et al. [97] | 2014 | Hybrid PSO with SA operator | Job-shop Scheduling |
| Sudibyo et al. [98] | 2015 | SA-PSO | Non-Linear Model Predictive Control |
| Wang and Sun [99] | 2016 | SA-PSO | K-Means Clustering |
| Javidrad and Nazari [100] | 2017 | PSO-SA | Global Optimization |
| Li et al. [102] | 2017 | SA-PSO | Parallel Plug-In Hybrid Electric Vehicle |

4.8.4. Hybridization of PSO using Ant Colony Optimization (ACO)

Ant Colony Optimization (ACO) proposed by Marco Dorigo [103] captured the organized communication triggered by an autocatalytic process practiced in ant colonies. In later years, Shelokar et al. [104] proposed PSACO (Particle Swarm Ant Colony Optimization) which implemented rapid global exploration of the search domain, while the local search was pheromone-guided. The first part of the algorithm works on PSO to generate initial solutions, while the positions of the particles are updated by ACO in the next part. This strategy proved to reach almost optimal solutions for highly non-convex problems. On the other hand, Kaveh et al. [105] introduced Discrete Heuristic Particle Swarm Ant Colony Optimization (DHPSACO) incorporating a fly-back mechanism [106]. It was concluded to be a fast algorithm with high convergence speed. Niknam and Amiri [107] combined a fuzzy adaptive Particle Swarm Optimization, Ant Colony Optimization and the K-Means algorithm for clustering analysis over a number of benchmark datasets and obtained improved performance in terms of good clustering partitions. They applied Q-learning, a reinforcement learning technique to ACO to come up with Hybrid FAPSO-ACO-K Algorithm. They compared the results with respect to PSO-ACO, PSO, SA, TS, GA, ACO, HBMO, PSO-SA, ACO-SA, K-Means and obtained better convergence of FAPSO-ACO-K in most cases provided the number of clusters known beforehand.

Chen et al. [108] proposed a Genetic Simulated Annealing Ant Colony system infused with Particle Swarm Optimization. The initial population of Genetic Algorithms was given by ACO, where the interaction among different groups about the pheromone information was controlled by PSO. Next, GA controlled by SA mutation techniques were used to produce superior results. Xiong and Wang [109] used a two-stage hybrid algorithm (TAPC) combining adaptive ACO and enhanced PSO to overcome the local optima convergence problem in a K-means clustering application environment. Kıran et al. [110] came up with a novel hybrid approach (HAP) combining ACO and PSO. While initially the individual behavior of randomly allocated swarm of particles and colony of ants gets predominance, they start getting influenced by each other through the global best solution, which is determined by comparing the best solutions of PSO and ACO at each iteration. Huang et al [111] introduced continuous Ant Colony Optimization (ACOR) in PSO to develop hybridization strategies based on four approaches out of which a sequence based approach using an enlarged pheromone-particle table proved to be most effective. The opportunity for ACOR in exploring the search space is more as solutions generated by PSO is associated with a pheromone-particle table. Mahi et al [112] came up with a hybrid approach combining PSO, ACO and 3-opt algorithm where the parameters concerning the optimization of ACO are determined by PSO and the 3-opt algorithm helps ACO to avoid stagnation in local optima.

Kefi et al. [113] proposed Ant-Supervised PSO (ASPSO) and applied it to the Travelling Salesman Problem (TSP) where the optimum values of the ACO parameters α and β, which used to determine the effect of pheromone information over the heuristic one, are updated by PSO instead of being constant as in traditional ACO. The pheromone amount and the rate of evaporation are detected by PSO: thus with the set of supervised and adjusted parameters given by PSO, ACO plays the key optimization methodology. Lazzus et al. [114] demonstrated vapor-liquid phase equilibrium by combining similar attributes of PSO and ACO (PSO+ACO), where the positions discovered by the particles of PSO were fine-tuned by the ants in the second stage through pheromone-guided techniques.

Mandloi et al. [115] presented a hybrid algorithm with a novel probabilistic search method by integrating the distance oriented search approach practiced by ants in ACO and velocity oriented search mechanism adopted by particles in PSO, thereby substituting the pheromone update of ACO with velocity update of PSO. The probability metric used in this algorithm consists of weighted heuristic values obtained from transformed distance and velocity through a sigmoid function, ensuring fast convergence, less complexity and avoidance of stagnation in local optima. Indadul et al. [116] solved the Travelling Salesman Problem (TSP) coordinating PSO, ACO and K-Opt Algorithm where the preliminary set of particle swarm is produced by ACO. In the latter iterations of PSO if the position of a particle is not changed for a given interval, then K-Opt Algorithm is applied to it for upgrading the position. Liu et al. [117] relied on the local search capacity of ACO and global search potential of PSO and conglomerated them for application in optimizing container truck routes. Junliang et al. [118] proposed a Hybrid Optimization Algorithm (HOA) that exploits the merit of global search and fast convergence in PSO and in the event of premature convergence lets ACO take over. With its initial parameters set by PSO, the algorithm then converges to the optimal solution.

A brief listing of some of the important hybrid algorithms using PSO and ACO are given below in Table 4.

Table 4. A collection of hybridized PSO-ACO algorithms.

| Author/s: | Year | Algorithm | Area of Application |
|---|---|---|---|
| Shelokar et al. [104] | 2007 | PSACO | Improved continuous optimization |
| Kaveh and Talatahari [105] | 2009 | DHPSACO | Truss structures with discrete variables |
| Kaveh and Talatahari [106] | 2009 | HPSACO | Truss structures |
| Niknam and Amiri [107] | 2010 | FAPSO-ACO-K | Cluster analysis |
| Chen et al. [108] | 2011 | ACO and PSO | Traveling salesman problem |
| Xiong and Wang [109] | 2011 | TAPC | Hybrid Clustering |
| Kıran et al. [110] | 2012 | HAP | Energy demand of Turkey |
| Huang et al. [111] | 2013 | ACOR | Data clustering |
| Mahi et al. [112] | 2015 | PSO, ACO and 3-opt algorithm | Traveling salesman problem |
| Kefi et al. [113] | 2015 | ASPSO | Traveling salesman problem |
| Lazzus et al. [114] | 2016 | PSO+ACO | Interaction parameters on phase equilibria |
| Mandloi and Bhatia [115] | 2016 | PSO, ACO | Large-MIMO detection |
| Indadul et al. [116] | 2017 | PSO, ACO and K-Opt Algorithm | Traveling salesman problem |
| Liu et al. [117] | 2017 | PSO, ACO | Container Truck Route optimization |
| Junliang et al. [118] | 2017 | HOA | Traveling salesman problem |

4.8.5. Hybridization of PSO using Cuckoo Search (CS)

Cuckoo Search (CS) proposed by Xin-She Yang and Suash Deb [119] was developed on the basis of breeding behavior of cuckoos associated with the levy flight nature of birds and flies.

Subsequently, Ghodrati and Lotfi [120] introduced Hybrid CS/PSO Algorithm capturing the ability of the cuckoos to communicate with each other in order to decrease the chances of their eggs being identified and abandoned by the host birds, by using Particle Swarm Optimization (PSO). In the course of migration each cuckoo records its personal best, thus generating the global best and governing their movements accordingly. Nawi et al. [121] came up with Hybrid Accelerated Cuckoo Particle Swarm Optimization (HACPSO) where the initial population of the nest are given by CS whereas Accelerated PSO (APSO) guides the agents towards the solution of the best nest. HACPSO was shown to perform classification problems with fast convergence and improved accuracy over constituent algorithms.

Enireddy and Kumar [122] proposed a hybrid PSO-CS for optimizing neural network learning rates. The meta parameter optimization of CS, i.e. the optimal values of the parameters of CS governing the rate of convergence of the algorithm were obtained through PSO which was shown to guarantee faster learning rate of neural networks with enhanced classification accuracy. Ye et al. [123] incorporated CS with PSO in the optimization of Support Vector Machine (SVM) parameters used for classification and identification of peer-to-peer traffic. At the beginning of each iteration, the optimal positions generated by PSO serve as the initial positions for CS and the position vectors of CS-PSO are considered as the pair of candidate parameters of SVM. The algorithm aims at finding the optimal tuning parameters of SVM through calculating the best position vectors.

Li and Yin [124] proposed a PSO inspired Cuckoo Search (PSCS) to model the update strategy by incorporating neighborhood as well as best individuals, balancing the exploitation and exploration capability of the algorithm. Chen et al. [125] combined the social communication of PSO and searching ability of CS and proposed PSOCS where cuckoos close to good solutions communicate with each other and move slowly near the optimal solutions guided by the global bests in PSO. This algorithm was used in training feedforward neural networks. Guo et al. [126] mitigated the shortcoming of PSO of getting trapped into local optima in high dimensional intricate problems through exploiting the random Levy step size update feature of CS, thus strengthening the global search ability. Also to overcome the slower convergence and lower accuracy of CS, they proposed hybrid PSOCS which initially uses Levy flight mechanism to search and then directs the particles towards optimal configuration through updating the positions given by PSO ensuring avoidance of local optima with randomness involved in Levy flights, thus providing improved performance.

Chi et al. [127] came up with a hybrid algorithm CSPSO where the initial population was based on the principles of orthogonal Latin squares with dynamic step size update using Levy flight process. The global search capability of PSO has been exploited ensuring information exchange among the cuckoos in the search process. Dash et al. [128] introduced improved cuckoo search particle swarm optimization (ICSPSO) where the optimization strategy of Differential Evolution (DE) Algorithm is incorporated for searching effectively around the group best solution vectors at each iteration, ensuring the global search capability of hybrid CSPSO and implemented this in designing linear phase multi-band stop filters.

A brief listing of some of the important hybrid algorithms using PSO and CS are given below in Table 5.

**Table 5.** A collection of hybridized PSO-CS algorithms.

| Author/s: | Year | Algorithm | Area of Application |
|---|---|---|---|
| Ghodrati and Lotfi [120] | 2012 | Hybrid CS/PSO | Global optimization |
| Nawi et al. [121] | 2014 | HACPSO | Classification |
| Enireddy and Kumar [122] | 2015 | Hybrid PSO CS | Compressed image classification |
| Ye et al. [123] | 2015 | Hybrid CSA with PSO | Optimization of parameters of SVM |
| Li and Yin [124] | 2015 | PSCS | Global optimization |
| Chen et al. [125] | 2015 | PSOCS | Artificial Neural Networks |
| Guo et al. [126] | 2016 | PSOCS | Preventive maintenance period optimization model |
| Chi et al. [127] | 2017 | CSPSO | Optimization problems |
| Dash et al. [128] | 2017 | ICSPSO | Linear phase multi-band stop filters |

4.8.6. Hybridization of PSO using Artificial Bee Colony (ABC)

Artificial Bee Colony (ABC) was proposed by Karaboga and Basturk [129] and models the organized and distributed actions adapted by colonies of bees. Shi et al. [130] proposed an integrated algorithm based on ABC and PSO (IABAP) by parallelly executing ABC and PSO and exchanging information between the swarm of particles and colony of bees. El-Abd [131] combined ABC and Standard PSO (SPSO) to update the personal best in SPSO using ABC at each iteration and applied it in continuous function optimization. Kıran and Gündüz [132] came up with a hybrid approach based on Particle Swarm Optimization and Artificial Bee Colony algorithm (HPA) where at the end of each iteration, recombination of the best solutions obtained by PSO and ABC takes place and the result serves as global best for PSO and neighbor for the onlooker bees in ABC, thus enhancing the exploration-exploitation capability of the algorithm.

Xiang et al. [133] introduced a Particle Swarm inspired Multi-Elitist ABC algorithm (PS-MEABC) in order to enhance the exploitation strategy of ABC algorithm by modifying the food source parameters in on-looker or employed bees phase through the global best as well as an elitist selection from the elitist archive. Vitorino et al. [134] came up with a way to deal with the issue that PSO is not always able to employ its exploration and exploitation mechanism in a well-adjusted manner, by trying a mitigation approach using the diversifying capacity of ABC when the agents stagnate in the search region. As the particles stagnate, the ABC component in Adaptive PSO (APSO) introduces diversity and enables the swarm to balance exploration and exploitation quotients based on fuzzy rules contingent upon swarm diversity.

Lin and Hsieh [135] used Endocrine-based PSO (EPSO) compensating a particle's adaptability by supplying regulatory hormones controlling the diversity of the particle's displacement and scope of search and combined it with ABC. The preliminary food locations of employed bees phase in ABC is supplied by EPSO's individual bests controlled by global bests. Then onlooker bees and scout bees play an important role in improving the ultimate solution quality. Zhou and Yang [136] proposed PSO-DE-PABC and PSO-DE-GABC based on PSO, DE and ABC to cope with the lack of exploitation plaguing ABC. Divergence is enhanced by creating new positions surrounding random particles through PSO-DE-PABC, whereas PSO-DE-GABC creates the positions around global best with the divergence being taken care by differential vectors and Dimension Factor (DF) optimizing the rate of search. The novelty in the scouting technique enhances search at the local level.

Li et al. [137] introduced PS-ABC comprising a local search phase of PSO and two global search phases of ABC. Depending on the extent of aging of personal best in PSO each entity at each iteration adopts either of PSO, onlooker bee or scout bee phase. This algorithm was shown to be efficient for high dimensional datasets with faster convergence. Sedighizadeh and Mazaheripour [138] came up with a PSO-ABC algorithm with initially assigned personal bests for each entity and further refining it through a PSO phase and ABC phase and the best of all personal bests is returned as global best at the end. This algorithm was shown to find the optimal route faster in vehicle routing problems compared to some competing algorithms.

A brief listing of some of the important hybrid algorithms using PSO and ABC are given below in Table 6.

Table 6. A collection of hybridized PSO-ABC algorithms.

| Author/s: | Year | Algorithm | Area of Application |
|---|---|---|---|
| Shi, et al. [130] | 2010 | IABAP | Global optimization |
| El-Abd [131] | 2011 | ABC-SPSO | Continuous function optimization |
| Kıran and Gündüz [132] | 2013 | HPA | Continuous optimization problems |
| Xiang, et al. [133] | 2014 | PS-MEABC | Real parameter optimization |
| Vitorino, et al. [134] | 2015 | ABeePSO | Optimization problems |
| Lin and Hsieh [135] | 2015 | EPSO_ABC | Classification of Medical Datasets Using SVMs |

| Zhou and Yang [136] | 2015 | PSO-DE-PABC and PSO-DE-GABC | Optimization problems |
|---|---|---|---|
| Li, et al. [137] | 2015 | PS-ABC | High-dimensional optimization problems |
| Sedighizadeh and Mazaheripour [138] | 2017 | PSO-ABC | Multi objective vehicle routing problem |

4.8.7. Hybridization of PSO using Other Social Metaheuristic Approaches

The following section discusses some instances where the Particle Swarm Optimization algorithm has been hybridized with other commonly used social metaheuristic optimization algorithms for use in an array of engineering applications. Common techniques include Artificial Immune Systems [139-142], Bat Algorithm [143], Firefly Algorithm [144] and Glow Worm Swarm Optimization Algorithm [145].

4.8.7.1 Artificial Immune Systems (AIS)

Zhao et al. [146] introduced a human-computer cooperative PSO based Immune Algorithm (HCPSO-IA) for solving complex layout design problems. The initial population is supplied by the user and the initial algorithmic solutions are generated by a chaotic strategy. By introducing new artificial individuals to replace poor performing individuals of the population, HCPSO-IA can be refined to incorporate a man-machine synergy by using knowledge about which key performance indices such as envelope area and static non-equilibrium value can be significantly increased.

El-Shirbiny and Alhamali [147] used a Hybrid Particle Swarm with Artificial Immune Learning (HPSIL) to solve Fixed Charge Transportation Problems (FCTPs). In the proposed algorithm a flexible particle structure, decoding as well as allocation scheme are used instead of a Prufer number and a spanning tree is used in Genetic Algorithms. The authors noted that the allocation scheme guaranteed finding an optimal solution for each of the particles and that the HPSIL algorithm can be implemented on both balanced and unbalanced FCTPs while not introducing any dummy supplier or demand.

4.8.7.2 Bat Algorithm (BA)

A communication strategy of hybrid PSO with Bat Algorithm (BA) was proposed in [148] by Pan et al. wherein several worst performing particles of PSO are replaced by the best performing ones in BA and vice-versa, after executing a fixed number of iterations. This communication strategy facilitates information flow between PSO and BA and can reinforce the strengths of each algorithm between function evaluations. Six benchmark functions were tested producing improvements in convergence speed and accuracy over either of PSO or BA.
An application of a hybrid PSO-BA in medical image registration was demonstrated by Manoj et al [149] where the authors noted that the hybrid algorithm was more successful in finding optimal parameters for the problem as compared to relevant methods already in use.

4.8.7.3 Firefly Algorithm (FA)

To utilize different advantages of PSO and Firefly Algorithm (FA), Xia et al. [150] proposed three novel operators in a hybrid algorithm (FAPSO) based on the two. During the optimization, the population is divided into two groups and each choses PSO or FA as their basic search technique for parallel executions. Apart from this, the information exchange about optimal solutions in case of stagnation, a knowledge based detection operator and a local search for tradeoff between exploration and exploitation as well as the employment of a BFGS Quasi-Newton method to enhance exploitation led to several important observations. For instance, the exchange of optimal solutions led to an enriched and diverse population and the inclusion of the detection operator and local search was justifiable for multimodal optimization problems where refinement of solution quality was necessary.

Arunachalam et al. [151] presented a new approach to solve the Combined Economic and Emission Dispatch (CEED) problem having conflicting economic and emission objective using a hybrid of PSO and FA.

4.8.7.4 Glow Worm Swarm Optimization (GSO)

Shi et al. [152] introduced a hybrid PSO and Glow Worm Swarm (GSO) algorithm (HEPGO) based on selective ensemble learning and merits of PSO and GSO. HEPGO leverages the ability of the GSO to capture multiple peaks of multimodal functions due to its dynamic sub-groups and the global exploration power of the PSO and provides promising results on five benchmark minimization functions.

Liu and Zhou [153] introduced GSO in the working of PSO to determine a perception range, within the scope of perception of all particles to find an extreme value point sequence. A roulette wheel selection scheme for picking a particle as the global extreme value is followed and the authors note that this can overcome the convergence issues faced by PSO.

A brief listing of some of the important hybrid algorithms using PSO and other approaches such as AIS, BA, FA and GSO are given below in Table 7.

Table 7. A collection of PSO algorithms hybridized with other approaches such as AIS, BA, FA and GSO.

| Author/s: | Year | Algorithm | Area of Application |
|---|---|---|---|
| Shi et al. [152] | 2012 | HEPGO | Global Optimization |
| El-Shirbiny and Alhamali [147] | 2013 | HPSIL | Fixed Charge Transportation Problems |
| Liu and Zhou [153] | 2013 | New (GSO-PSO) | Constrained Optimization |
| Zhao et al [146] | 2014 | HCPSO-IA | Complex layout design problems |
| Arunachalam et al. [151] | 2014 | HPSOFF | Combined Economic and Emission Dispatch Problem |
| Pan et al. [148] | 2015 | Hybrid PSO-BA | Global Optimization |
| Manoj et al. [149] | 2016 | PSO-BA | Medical Image Registration |
| Xia et al. [150] | 2017 | FAPSO | Global Optimization |

*4.9. Parallelized Implementations of PSO*

The literature points to several instances of PSO implementations on parallel computing platforms. The use of multiple processing units onboard a single computer renders feasibility to the speedup of independent computations in the inherently parallel structure of PSO. Establishing sub-swarm based parallelism leads to different processors being assigned to sub-swarms with some mechanism of information exchange among them. On the other hand, master-slave configurations attempt to designate a master processor which assigns slave processors to work on fitness evaluation of many particles simultaneously. Early work by Gies and Rahmat-Samii reported a performance gain of 8 times using a system with 10 nodes for a parallel implementation over s serial one [154]. Schutte et al. [155] evaluated a parallel implementation of the algorithm on two types of test problems: a) large scale analytical problems with inexpensive function evaluations and b) medium scale problems on biomechanical system identification with computationally heavy function evaluations. The results of experimental analysis under load-balanced and load-imbalanced conditions highlighted several promising aspects of parallelization. The authors used a synchronous scheme based on a master-slave approach. The use of data pools in [156], independent evaluation of fitness leading to establishing the dependency of efficiency on the social information exchange strategy in [157] and exploration of enhanced topologies for information exchange in multiprocessor architectures in [158] may be of relevance to an interested reader.

Rymut's work on parallel PSO based Particle Filtering showed how CUDA capable Graphics Processing Units (GPUs) can accelerate object tracking algorithm performances using adaptive appearance models [159]. A speedup factor of 40 was achieved by using GPUs over CPUs. Zhang et al. fused GA and PSO to address the sample impoverishment problem and sample size dependency in particle filters [160-161]. Chen et al. [162] proposed an efficient parallel PSO algorithm to find optimal design criteria for Central Composite Discrepancy (CCD) criterion whereas Awwad used a CUDA based approach to solve the topology control issue in hybrid radio frequency and wireless networks in optics [163]. Qu et al. used a serial and parallel implementation of PSO in the Graph Drawing problem [164] and reported that both methods are as effective as the force-directed method in the work, with the parallel method being superior to the serial one when large graphs were considered. Zhou et al. found that using a CUDA implementation of the Standard PSO (SPSO) with a local topology [165] on four benchmark problems, the runtime of GPU-SPSO indicates clear superiority over CPU-SPSO. They also noted that runtime and swarm size assumed a linear relationship in case of GPU-SPSO. Mussi et al. reported in [166] an in-depth performance evaluation of two variants of parallel algorithms with the sequential implementation of PSO over standard benchmark functions. The study included assessing the computational efficiency of the parallel methods by considering speedup and scaleup against the sequential version.

## 5. Niche Formation and Multi-Objective Optimization

A function is multimodal if it has more than one optimum. Multimodal functions may have one global optimum with several local optima or more than one global optimum. In the latter case, optimization algorithms are refined appropriately to be effective on multimodal fitness landscapes as two specific circumstances may arise otherwise. First, standard algorithms used may be unable to distinguish among the promising regions and settle on a single optimum. Second, the algorithm may not converge to any optima at all. In both cases, the multi-objective optimization criteria are not satisfied and further modifications are needed.

*5.1 Formation of Niches in PSO*

The formation of niches in swarms is inspired by the natural phenomenon of co-existence of species who are competing and co-evolving for shared resources in a social setting. The work of Parsopoulous et al. [167-168] was among the first ones to appropriately modify PSO to make it suitable for handling multimodal functions with multiple local optima through the introduction of function "stretching" whereby fitness neighborhoods are adaptively modified to remove local optima. A sequential niching technique proposed by Parsopoulous and Vrahatis [169] identified possible solutions when their fitness dropped below a certain value and raised them, at the same time removing all local optima violating the threshold constraint. However, the effectiveness of the stretching method is not uniform on all objective functions and introduced false minima in some cases [33]. This approach was improved by the introduction of the *Deflection* and *Repulsion* techniques in [170] by Parsopoulous and Vrahatis. The nbest PSO by Brits et al. [171] used local neighborhoods based on spatial proximity and achieved a parallel niching effect in a swarm whereas the NichePSO by the same authors in [172] achieved multiple solutions to multimodal problems using sub-swarms generated from the main swarm when a possible niche was detected. A speciation-based PSO [173] was developed keeping in mind the classification of particles within a threshold radius from the neighborhood best (also known as the seed), as those belonging to a particular species. In an extension which sought to eliminate the need for a user specified radius of niching, the Adaptive Niching PSO (ANPSO) was proposed in [174-175]. It adaptively determines the radius by computing the average distance between each particle and its closest neighbor. A niche is said to have formed if particles are found to be within the niching radius for an extended number of iterations in which case particles are classified into two groups: niched and un-niched. A global PSO is used for information exchange within the niches whereas an lbest PSO with a Von-Neumann topology is used for the same in case of un-niched particles. Although ANPSO eliminates the requirement of specifying a niche radius beforehand, the solution quality may become sensitive due to the addition of new parameters. The

Fitness Euclidean Distance Ratio PSO (FER-PSO) proposed by Li [176] uses a *memory swarm* alongside an *explorer swarm* to guide particles towards the promising regions in the search space. The memory swarm is constructed out of personal bests found so far whereas the explorer swarm is constructed out of the current positions of the particle. Each particle is attracted towards a fittest and closest point in the neighborhood obtained by computing its Fitness Euclidean Ratio (FER). FER-PSO introduces a scaling parameter in the computation of FER, however it can reliably locate all global optima when population sizes are large. Clustering techniques such as k-means have been incorporated into the PSO framework by Kennedy [177] as well as by Passaro and Starita [178] who used the Bayesian Information Criterion (BIC) [179] to estimate the parameter k and found the approach comparable to SPSO and ANPSO.

*5.2. Niching in Dynamic Environments and Challenges*

Several hard challenges are posed by environments which are dynamic as well as multimodal, however sub-population based algorithms searching in parallel are an efficient way to locate multiple optima which may undergo any of shape, height or depth changes as well as spatial displacement. Multi-Swarm PSO proposed by Blackwell et al. [180], rPSO by Bird and Li [181], Dynamic SPSO by Parrot and Li [182] and the *lbest* PSO with Ring Topology by Li [183] are some of the well-known approaches used in such environments. Since most niching algorithms utilize global information exchange at some point in their execution, their best case computational complexity is $O(N^2)$. This issue coupled with performance degradation in high dimensional problems and the parameter sensitivity of solutions make niching techniques an involved process for any sufficiently complex multimodal optimization problem.

## 6. Discrete Hyperspace Optimization

6.1.1. Variable Round-Off

Discrete variables are rounded off to their nearest values by clamping at the end of every iteration or at the end of the optimization process and can offer significant speedup. However, unintelligent round-offs may throw the particle towards a comparably infeasible region and result worse fitness values. With that said, some studies have shown interesting results and garnered attention for a discrete version of PSO (DPSO).

6.1.2. Binarization

A widely used binarization approach maps the updated velocity at the end of an iteration into the closed interval [0,1] using a sigmoid function. The updated velocity represents the probability that the updated position takes the value of 1 since a high enough velocity implies the sigmoid function outputs 1. The value of the maximum velocity is often clamped to a low value to make sure there is a chance of a reversal of sigmoid output value. Afshinmanesh et al. [184] used modified flight equations based on XOR and OR operations in Boolean algebra. Negative selection mechanisms in immune systems inspire a velocity bounding constraint on such approaches. Deligkaris et al [185] used a mutation operator on particle velocities to render better exploration capabilities to the Binary PSO.

6.1.3. Set Theoretic Approaches

Chen et al. [186] used a set representation approach to characterize the discrete search spaces of combinatorial optimization problems. The solution is represented as a crisp set and the velocity as a set of possibilities. The conventional operators in position and velocity update equations of PSO are replaced by operators defined on crisp sets and sets of possibilities, thus enabling a structure similar to PSO but with applicability to a discrete search space. Experiments on two well-known discrete optimization problems, viz. the Traveling Salesman Problem (TSP) and the Multidimensional Knapsack Problem (MKP) demonstrated the promising nature of the discrete version. Gong et al.

[187] proposed a set-based PSO to solve the Vehicle Routing Problem (VRP) with Time Windows (S-PSO-VRPTW) with the general method of selecting an optimal subset from a universal set using the PSO framework. S-PSO-VRPTW considers the discrete search space as an arc set of the complete graph represented by the nodes in VRPTW and regards a potential solution as a subset of arcs. The designed algorithm when tested on Solomon's datasets [188] yielded superior results in comparison to existing state-of-the-art methodologies.

6.1.4. Penalty Approaches

KitaYama et al. [189] setup an augmented objective function with a penalty approach such that there is a higher incentive around discrete values. Points away from discrete values are penalized and the swarm effectively explores a discrete search space, although at a heavy computational burden for complex optimization problems. By incorporating the penalty term the augmented objective function turns non-convex and continuous, thereby making it suitable for a PSO based optimization.

6.1.5. Hybrid Approaches

Nema et al. [190] used the deterministic Branch and Bound algorithm to hybridize PSO in order to solve the Mixed Discrete Non-Linear Programming (MDNLP) problem. The global search capability of PSO coupled with the fast convergence rate of Branch and Bound reduces the computational effort required in Non-Linear Programming problems. Sun et al. [191] introduced a constraint preserving mechanism in PSO (CPMPSO) to solve mixed-variable optimization problems and reported competitive results when tested on two real-world mixed-variable optimization problems. Chowdhury et al. [192] considered the issue of premature stagnation of candidate solutions especially in single objective, constrained problems when using PSO and noted its pronounced effect in objective functions that make use of a mixture of continuous and discrete design variables. In order to address this issue, the authors proposed a modification in PSO which made use of continuous optimization as its primary strategy and subsequently a nearest vertex approximation criterion for updating of discrete variables. Further incorporation of a diversity preserving mechanism introduced a dynamic repulsion directed towards the global best in case of continuous variables and a stochastic update in case of discrete ones. Performance validation tests were successfully carried out over a set of nine unconstrained problems and a set of 98 Mixed Integer Non-Linear Programming (MINLP) problems.

6.1.6. Some application instances

Laskari et al [193] tested three variants of PSO against the popular Branch and Bound method on seven different integer programming problems. Experimental results indicated that the behavior of PSO was stable in high dimensional problems and in cases where Branch and Bound failed. The variant of PSO using constriction factor and inertia weight was the fastest whereas the other variants possessed better global exploration capabilities. Further observation also supported the claim that the variant of PSO with only constriction factor was significantly faster than the one with only inertia weight. These results affirmed that the performance of the variants was not affected by truncation of the real parameter values of the particles. Yare and Venayagamoorthy [194] used a discrete PSO for optimal scheduling of generator maintenance, Eajal and El-Hawary [195] approached the problem of optimal placement and sizing of capacitors in unbalanced distribution systems with the consideration of including harmonics. More recently, Phung et al. [196] used a discretized version of PSO path planning for UAV vision-based surface inspection and Gong et al. [197] attempted influence maximization in social networks. Aminbakhsh and Sonmez [198] presented a Discrete Particle Swarm Optimization (DPSO) for an effective solution to large-scale Discrete Time-Cost Trade-off Problem (DTCTP). The authors noted that the experiments provided high quality solutions for time-cost optimization of large size projects within seconds and enabled optimal planning of real life-size projects. Li et al. [199] modeled complex network clustering as a multi-objective optimization problem and applied a quantum inspired discrete particle swarm optimization algorithm with non-dominated sorting for individual replacement to solve it. Experimental results illustrated its competitiveness against some state-of-the-art approaches on the extensions of Girvan and Newman benchmarks [200] as well as many real-world networks. Ates et al. [201] presented a discrete Infinite Impulse Response (IIR) filter design method for approximate realization of fractional order continuous filters using a Fractional Order Darwinian Particle Swarm Optimization (FODPSO).

## 7. Ensemble Particle Swarm Optimization

The No Free Lunch (NFL) Theorem [202] by Wolpert and Macready establishes that no single optimization algorithm can produce superior results when averaged over all objective functions. Instead, different algorithms perform with different degrees of effectiveness given an optimization problem. To this end, researchers have tried to put together ensembles of optimizers to obtain a set of candidate solutions given an objective function and choose from the promising ones.

Existing ensemble approaches such as the Multi-Strategy Ensemble PSO (MEPSO) [203] uses a two-stage approach: Gaussian local search strategy to improve convergence capability and Differential Mutation (DM) to increase the diversity of the particles. The Heterogenous PSO in [204] uses a pool of different search behaviors of PSO and empirically outperforms the homogenous version of PSO. The Ensemble Particle Swarm Optimizer by Lynn and Suganthan [205] uses a pool of PSO strategies and gradually chooses a suitable one through a merit-based scheme to guide the particles' movement in a particular iteration.

Shirazi et al. proposed a Particle Swarm Optimizer with an ensemble of inertia weights in [206] and tested its effectiveness by incorporating it into a heterogenous comprehensive learning PSO (HCLPSO) [207]. Different strategies such as linear, logarithmic, exponential decreasing, Gompertz, chaotic and oscillating inertia weights were considered and compared against other strategies on a large set of benchmark problems with varying dimensions to demonstrate the suitability of the proposed algorithm.

## 8. Notes on Benchmark Solution Quality and Performance Comparison Practices

### 8.1. Performance on Simple Benchmarks

To provide an intuitive understanding of the performances of some of the many PSO-based algorithmic variants, let us consider a few commonly used unimodal and multimodal benchmark functions. These functions (F1-F8) are either unimodal, simple multimodal or unrotated multimodal ones. The purpose of the following table is to provide a first-course reference for introductory purposes, however for a full-scale performance analysis one should consider rotated multimodal and compositional functions as well – there should be a good mix of separable and non-separable benchmarks before any inference on accuracy and/or efficiency is drawn.

**Table 8. Benchmark Functions F1-F8**

| Function | Name | Expression | Range | Min |
|---|---|---|---|---|
| F1 | Sphere | $f(x) = \sum_{i=1}^{n} x_i^2$ | [-100,100] | f(x*) = 0 |
| F2 | Schwefel's Problem 2.22 | $f(x) = \sum_{i=1}^{n} |x_i| + \prod_{i=1}^{n} |x_i|$ | [-10,10] | f(x*) = 0 |
| F3 | Schwefel's Problem 1.2 | $f(x) = \sum_{i=1}^{n} \left( \sum_{j=1}^{i} x_j \right)^2$ | [-100,100] | f(x*) = 0 |
| F4 | Generalized Rosenbrock's Function | $f(x) = \sum_{i=1}^{n-1} [100(x_{i+1} - x_i^2)^2 + (x_i - 1)^2]$ | [-n,n] | f(x*) = 0 |
| F5 | Generalized Schwefel's Problem 2.26 | $f(x) = -\sum_{i=1}^{n} (x_i \sin(\sqrt{|x_i|}))$ | [-500,500] | f(x*) = -12569.5 |

| | | | | | |
|---|---|---|---|---|---|
| F6 | Generalized Rastrigrin's Function | $f(x) = An + \sum_{i=1}^{n}[x_i^2 - A\cos(2\pi x_i)]$, A=10 | [-5.12, 5.12] | f(x*) = 0 |
| F7 | Ackley's Function | $f(x) = -20\exp\left(-0.2\sqrt{\frac{1}{d}\sum_{i=1}^{d}x_i^2}\right) - \exp\left(\sqrt{\frac{1}{d}\sum_{i=1}^{d}\cos(2\pi x_i)}\right) + 20 + \exp(1)$ | [-32.768, 32.768] | f(x*) = 0 |
| F8 | Generalized Griewank Function | $f(x) = 1 + \frac{1}{4000}\sum_{i=1}^{n}x_i^2 - \prod_{i=1}^{n}\cos(\frac{x_i}{\sqrt{i}})$ | [-600,600] | f(x*) = 0 |

**Table 9. 3D Plots of the Benchmark Functions**

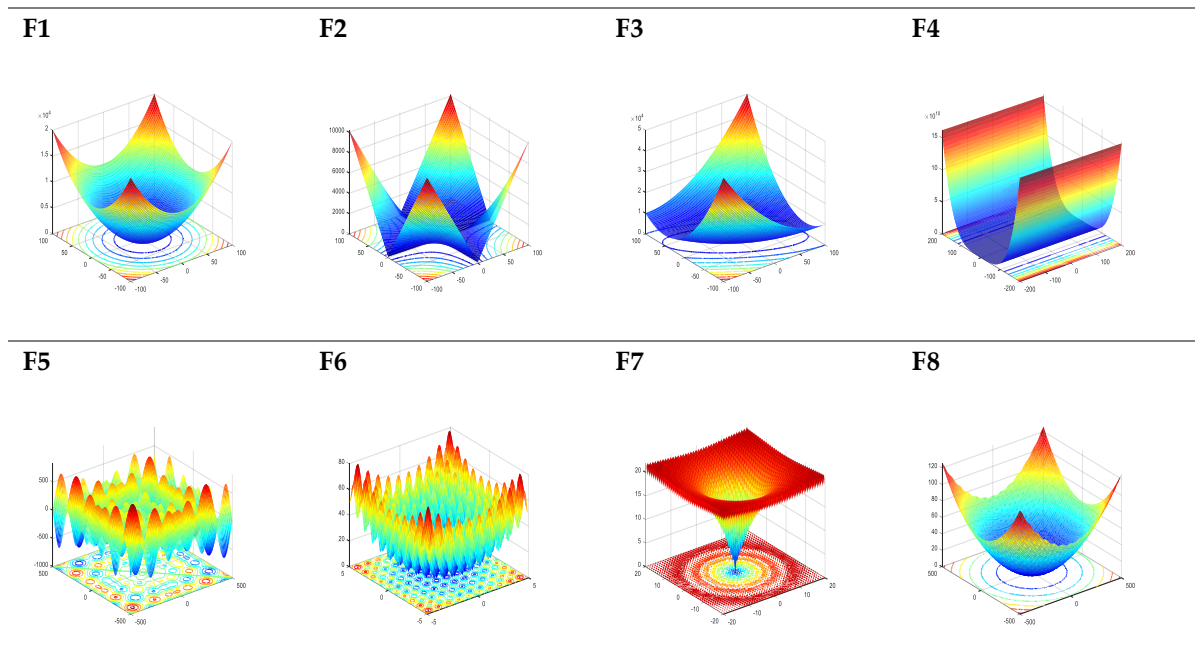

| F1 | F2 | F3 | F4 |
| F5 | F6 | F7 | F8 |

**Table 10. Performances of Some Variants of PSO on F1-F8**

| Function | Performance | PSO [208] | PSO [209] | PSO [210] | PSOGSA [210] | DEPSO [211] |
|---|---|---|---|---|---|---|
| F1 | Mean | 1.36e-04 | 1.8e-03 | 2.83e-04 | 6.66e-19 | 1.60e−26 |
|  | St. Dev | 2.02e-04 | NR | NR | NR | 6.56e−26 |
| F2 | Mean | 4.21e-02 | 2.0e+0 | 5.50e-03 | 3.79e-09 | 2.89e−13 |
|  | St. Dev | 4.54e-02 | NR | NR | NR | 1.54e−12 |
| F3 | Mean | 7.01e+01 | 4.1e+03 | 5.19e+3 | 4.09e+02 | 3.71e−01 |
|  | St. Dev | 2.21e+01 | NR | NR | NR | 2.39e−01 |
| F4 | Mean | 9.67e+01 | 3.6e+04 | 2.01e+02 | 5.62e+01 | 4.20e+01 |
|  | St. Dev | 6.01e+01 | NR | NR | NR | 3.28e+01 |
| F5 | Mean | -4.84e+03 | -9.8e+03 | -5.92e+03 | -1.22e+04 | 4.68e+03 |
| | St. Dev | 1.15e+03 | NR | NR | NR | 9.42e+02 |
| F6 | Mean | 4.67e+01 | 5.51e+01 | 7.23e+01 | 2.27e+01 | 4.07e+01 |
| | St. Dev | 1.16e+01 | NR | NR | NR | 1.19e+01 |
| F7 | Mean | 2.76e-01 | 9.0e-03 | 4.85e-10 | 6.68e-12 | 2.98e–13 |
| | St. Dev | 5.09e-01 | NR | NR | NR | 1.51e–12 |
| F8 | Mean | 9.21e-03 | 1.0e-02 | 5.43e-03 | 1.48e-03 | 1.69e–02 |
| | St. Dev | 7.72e-03 | NR | NR | NR | 1.82e–02 |

*NR – Not Reported

The results in Table 10 provide a high-level, intuitive understanding of benchmarking experiments using PSO and a few of its variants. In order to comment on the performance of the algorithms, statistical significance tests with an appropriate confidence level (generally alpha=0.01 or 0.05) are commonly carried out.

8.1. Studies on Performance Comparison Practices

Sergeyev et al. [212] proposed a visual technique for comparison of different approaches towards global optimization problems from both stochastic point of view with metaheuristics inspired by nature as well as deterministic approach based on mathematical programming. They have presented proposed and aggregated operational zones for effective comparison of deterministic and stochastic algorithms for various computational budgets. They commented on the competitive nature of the two algorithms and the fact that they surpass each other based on the available cost function evaluations. However, one shortcoming of this approach is the apparent under-performance of an algorithm with respect to its potential, given that it stagnates in a local minimum before the users' computational budget is exhausted. In order to work around this the authors put forward two budgetary upper limits viz. $n_{max}$ and $N_{max}$ (typically, $N_{max} \gg n_{max}$). The underlying strategy is to let any algorithm operate on any objective function within an upper limit of $n_{max}$. It is only necessary to check if the function is approximated within the global budget $N_{max}$ or if a successful trial is reported conditioned upon success in either an individual trial with a maximum local budget of $n_{max}$ as well as in a batch of trials each with the same local budget of $n_{max}$. Post-optimization data is used to construct *aggregate operational zones*. This approach relies on different reinitializations across $k$ different local budgets with the aim of maximizing an exploration-exploitation gain, while keeping the global budget constant. It is important to note that different trials may require different runtimes to converge and that the condition to check if an objective function is *successfully* approximated is rather flexible.

Kvasov and Mukhametzhanov [213-214] considered the problem of finding the global optimum $f^*$ and the corresponding argument $x^*$ of continuous and finite-dimensional objective functions (specifically, unidimensional ones) of multimodal, non-differentiable nature from a constrained optimization standpoint. Testing was carried out on 134 multimodal, constrained functions of univariate nature with respect to various performance comparison indices, totaling over 125,000 trials using 13 test methods. The experimental results provide critical insight into the comparative efficiencies of Lipschitz-based deterministic approaches versus nature-inspired metaheuristic ones, with future directions pointed at an extension of similar analyses for the multidimensional case.

Readers are directed to [213] for an involved understanding of the test methodology and to [215-216] for comparison criteria for application and validation in some practical test problems.

## 9. Future directions

Two decades of exciting developments in the Particle Swarm paradigm has seen many exciting upheavals and successes alike. The task of detecting a global among the presence of many local optima, the arbitrary nature of the search space and the intractability of using conventional mathematical abstractions on a wide range of objective functions coupled with little or no guarantees apriori about any optima being found made the search process challenging. However, Particle Swarm Optimizers have had their fair share of success stories – they can be used on any objective function: continuous or discontinuous, tractable or intractable, even for those where initialization renders solution quality to be sensitive as evidenced in case of their deterministic counterparts. However, some pressing issues which are listed below merit further work by the PSO community.

1. Parameter Sensitivity: Solution quality of metaheuristics like PSO are sensitive to their parametric evolutions. This means that the same strategy of parameter selection does not work for every problem.
2. Convergence to local optima: Unless the basic PSO is substantially modified to take into account the modalities of the objective function, more often than not it falls prey to local optima in the search space for sufficiently complex objective functions.
3. Subpar performance in multi-objective optimization for high dimensional problems: Although niching techniques render acceptable solutions for multimodal functions in both static and dynamic environments, the solution quality falls sharply when the dimensionality of the problem increases.

Ensemble optimizers, although promising, do not address the underlying shortcomings of the basic PSO. Theoretical issues such as the particle explosion problem, loss of particle diversity as well as stagnation to local optima deserve the attention of researchers so that a unified algorithmic framework with more intelligent self-adaptation and less user-specified customizations can be realized for future applications.

**Acknowledgments:** This work was made possible by the financial and computing support by the Vanderbilt University Department of EECS. The authors would like to thank the anonymous reviewers for their valuable comments for further improving the content of this article.

**Conflict of Interest:** The authors declare no conflict of interest.